# Holistic Filter Pruning for Efficient Deep Neural Networks

Lukas Enderich
Corporate Research
Robert Bosch GmbH, Renningen, Germany
lukas.enderich@bosch.com

Fabian Timm
Corporate Research
Robert Bosch GmbH, Renningen, Germany
fabian.timm@bosch.com

Wolfram Burgard
Institute for Autonomous Intelligent Systems
University of Freiburg, Germany
burgard@informatik.uni-freiburg.de

## Abstract

*Deep neural networks (DNNs) are usually over-parameterized to increase the likelihood of getting adequate initial weights by random initialization. Consequently, trained DNNs have many redundancies which can be pruned from the model to reduce complexity and improve the ability to generalize. Structural sparsity, as achieved by filter pruning, directly reduces the tensor sizes of weights and activations and is thus particularly effective for reducing complexity. We propose Holistic Filter Pruning (HFP), a novel approach for common DNN training that is easy to implement and enables to specify accurate pruning rates for the number of both parameters and multiplications. After each forward pass, the current model complexity is calculated and compared to the desired target size. By gradient descent, a global solution can be found that allocates the pruning budget over the individual layers such that the desired target size is fulfilled. In various experiments, we give insights into the training and achieve state-of-the-art performance on CIFAR-10 and ImageNet (HFP prunes 60% of the multiplications of ResNet-50 on ImageNet with no significant loss in the accuracy). We believe our simple and powerful pruning approach to constitute a valuable contribution for users of DNNs in low-cost applications.*

## 1. Introduction

Deep neural networks (DNNs) have a strong ability for data abstraction and outperform classical methods in many machine learning challenges such as computer vision, object detection, or speech recognition [1, 19]. But, recent progress has been made by training powerful models with many parameters using large scale data sets [7, 32]. Frankle and Carbin [3] demonstrated the correlation between the initial model size and the probability of getting meaningful initial values for the parameters by random initialization. As a result, modern DNNs are usually over-parameterized, have high memory requirements and need many floating-point multiplications, which are especially expensive concerning computation time and energy consumption [32].

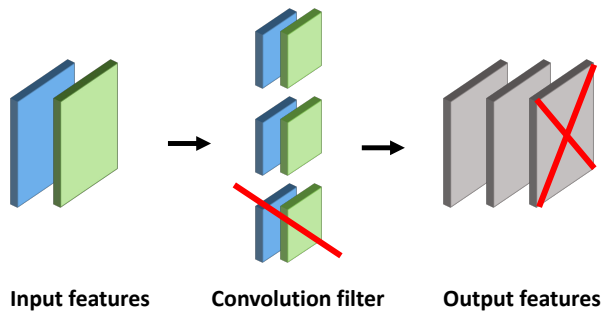

Figure 1. Structural sparsity can be achieved by pruning complete filters or neurons from the network. Since filter pruning reduces both the number of filters in the respective layer and the number of output feature maps, the tensor sizes of both weights and activations decrease. With a reduced number of output feature maps, the depth of the following layer decreases to the same degree.

However, reduction techniques can significantly reduce the complexity of trained DNNs. On the one hand, quantization methods reduce the precision of both parameters and activations to accelerate DNNs on dedicated hardware [32, 2]. On the other hand, pruning and factorization methods reduce the number of parameters and multiplications rather than their bit-sizes [22, 21]. Structural sparsity, as achieved by filter pruning, directly reduces computation time, energy consumption, and memory requirements without the need for specialized hardware. A visualization of filter pruning is given in figure 1.

Unsupervised filter pruning usually fails to preserve the accuracy of the original model [29]. Therefore, data driven approaches have been developed which either iteratively prune filters based on saliency scores [13, 22, 9, 38, 10, 24] or retrain the model under consideration of sparsity constraints [37, 11, 27, 31, 33]. Methods of the first category calculate saliency scores to rate the importance of individual filters. Filters with low saliency scores are considered unimportant and are deleted whereas the remaining filters are retrained. This process is repeated until the desired pruning rate is reached. In contrast, methods of the second category investigate sparsity constraints that can be integrated into the the training of DNNs. Regularization terms push the sum of absolute values of filter weights towards zero [27, 37, 11]. Furthermore, in [31, 33, 34] learnable gate variables were introduced that scale single weights or complete filters by one or zero.

However, most recent approaches have some disadvantages. Determining saliency scores requires a lot of human labor and is usually a heuristic practice. Furthermore, layer-by-layer pruning as well as iterative pruning and retraining are unsuitable procedures for determining a global selection of filters to be pruned. Considering that all networks layers jointly contribute to the learning task, it is inappropriate to prune single layers independently. Moreover, iterative pruning and retraining may prune filters that become important again at a later iteration.

In this work, we make the following contributions:

- We propose a holistic pruning approach that can be integrated into the training of DNNs by only a few lines of code. The proposed method induces sparsity via the channel-wise scaling factors of the batch-normalization layers. Hence, no additional variables are needed. Furthermore, the user can specify the desired model size in terms of the number of parameters and multiplications. The pruning budget is allocated over all layers automatically such that the desired model size is reached.

- We evaluate our pruning approach on two benchmark data-sets (CIFAR-10, ImgaeNet). We provide comparisons with recent filter pruning results and prove state-of-the-art performance on various DNN architectures. Furthermore, we analyze the allocation of pruning rates over the individual layers for different target sizes and layer types.

## 2. Related Work

DNNs are usually over-parameterized and have many redundant network connections, which can be eliminated (e.g. pruned) after the training to reduce the model complexity and improve the ability to generalize. The first pruning methods were aimed at setting single weights to zero in order to trim intermediate layer connections. *Optimal Brain Damage* [20] utilized the second-order derivatives of the loss function to calculate saliencies for each network weight. Subsequently, weights with small saliency scores were pruned iteratively whereas the remaining weights were retrained. Since calculating the second-order derivatives of the loss function with respect to the prameters is too complex for large DNNs, many approaches applied magnitude based pruning [6, 5, 27, 37].

However, since pruning single weights has no direct benefit for the hardware implementation of DNNs (unstructural sparsity), the indicator of non-zero weights is an insufficient evidence of the model complexity. In contrast, pruning complete filters or neurons from the network architecture directly reduces the tensor sizes of weights and activations (structural sparsity). A visualization is given in figure 1. Filter pruning methods can be devided into two subcategories[21]: *saliency based pruning and retraining* on the one hand and *sparsity learning* on the other. Both subcategories are based on pre-trained and usually over-parameterized models. In the following, a filter is equivalent to a channel or a neuron.

### 2.1. Saliency based pruning and retraining

These methods determine heuristics to calculate saliency scores for each filter. The saliency score indicates the importance of the respective filter: The higher the score the more important the filter is considered to be for fulfilling the learning task. Based on the saliencies, a certain number or percentage of filters is pruned whereas the remaining ones are retrained. This process is repeated iteratively until the desired network size is reached.

Hu *et al*. identified unimportant filters by analyzing the magnitudes of the output activations [13]. Feature maps with comparatively small sums of absolute values were considered less important and hence removed. In contrast, Li *et al*. measured the importance of individual filters by calculating the sum over the absolute values of the weights [22]. Zhuang *et al*. argued that informative channels should have discriminative power [38]: They proposed a ranking heuristic to identify channels with high discriminative power while deleting redundant channels and their corresponding filters. Furthermore, He *et al*. set the weights of filters with small $L^2$-norms to zero [9]. During the retraining steps, however, the pruned filters were updated as well to improve the training behaviour. The procedure is repeated until the selection of filters with small $L^2$-norms converges. Furthermore, Yu *et al*. calculated saliency scores by minimizing the reconstruction error in the second-to-last layer before the classification output [35]. Recently, Zhonghui et al introduced *Gate Decorator*, a pruning framework that uses gates to scale the channel-wise output of intermediate layers [34]. The change in the loss function caused by setting the gates to zero is calculated using a Tay-lor expansion and subsequently used for the saliency scores.

## 2.2. Sparsity learning

Sparsity learning induces sparsity constraints into the training of DNNs. Pan et al. approximated the $L^0$-norm to penalize incoming and outgoing connections of single filters [30]. He *et al.* proposed a channel selection based on LASSO regression whereas pruning each layer is achieved by minimizing the reconstruction error of the output feature maps [11]. Furthermore, Liu *et al.* applied $L^1$-norm based regularization on the scaling factors of the batch-normalization layers to scale single channels towards zero [26]. Subsequently, a certain percentile of channels is pruned according to a global threshold across all layers. However, in extreme cases this could lead to all channels of a single layer being pruned. Aditionally, the scaling factors are penalized without considering the respective filter size. In contrast, Huang *et al.* proposed a try-and-learn algorithm to train pruning agents that identify superfluous filters [15].

Recently, Xiao *et al.* introduced *Auto Prune*, a framework that uses a set of additional parameters to prune single weights or filters during each forward pass [33]. However, in their implementation, the pruning layers are located in front of the batch-normalization layers which reactivate the pruned channels (unless batch-normalization is disabled). Srinivas *et al.* proposed a similar approach using gate variables but neglected batch-normalization layers as well [31].

## 3. Background on batch-normalization

DNNs consist of interconnected layers which mainly perform weighted sums (convolution and fully-connected layers), batch-normalization, and non-linear transformations. With $l$ being the layer index, the weighted sums can be written as

$$a_l = w_l * x_{l-1} + b_l \quad (1)$$

with the layer input $x_{l-1}$, the weight-tensor or -matrix $w_l$, the bias vector $b_l$ and $*$ denoting either a convolution operator or a matrix-vector multiplication. Each layer consists of several channels, with the amount of channels in $a_l$ being equal to the number of convolution filters or matrix rows in $w_l$. After calculating the weighted sums, each channel is normalized and transformed linearly. The normalized output $\hat{a}_{l,c}$ is calculated by

$$\hat{a}_{l,c} = \begin{cases} \dfrac{a_{l,c} - \mathrm{E}[a_{l,c}]}{\sqrt{\mathrm{Var}[a_{l,c}] + \epsilon}} \gamma_{l,c} + \beta_{l,c} & \text{during training,} \\ \dfrac{a_{l,c} - \mu_{l,c}}{\sqrt{\sigma_{l,c}^2 + \epsilon}} \gamma_{l,c} + \beta_{l,c} & \text{during inference,} \end{cases} \quad (2)$$

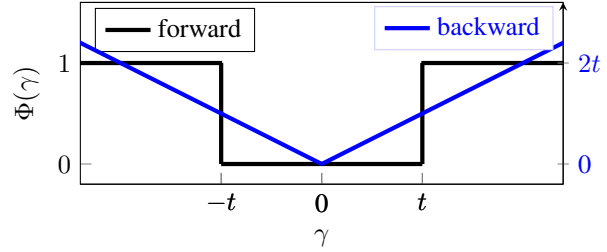

Figure 2. An illustration of the indicator function during both forward and backward pass. During the forward pass, the indicator function outputs whether the absolute values of the batch-normalization scaling factors are greater than $t$. During the backward pass, the indicator function is approximated using two piecewise linear functions. Thus, the gradient with respect to the scaling factors is either $\pm 1$, depending on the sign of the scaling factors.

with $c$ denoting the channel index, $\mathrm{E}[a_{l,c}]$ and $\mathrm{Var}[a_{l,c}]$ being the mean and the variance of the current mini-batch, and $\{\gamma_{l,c}, \beta_{l,c}\}$ being the learnable parameters of the affine transformation.

After training, batch-normalization layers are folded into the preceding convolution or fully-connected layer to accelerate the inference graph. The normalized output of the folded layer $\hat{a}_l$ can therefore be written as

$$\hat{a}_l = \hat{w}_l * x_{l-1} + \hat{b}_l \quad \text{with} \quad \hat{w}_l = w_l \frac{\gamma_l}{\sqrt{\sigma_l^2 + \epsilon}}$$

$$\text{and} \quad \hat{b}_l = (b_l - \mu_l) \frac{\gamma_l}{\sqrt{\sigma_l^2 + \epsilon}} + \beta_l. \quad (3)$$

Thus, batch-normalization scaling factors can be used to prune complete filters from the network structure: As the absolute value decreases, $\gamma_{l,c}$ scales the output of channel $c$ in layer $l$ towards zero.

## 4. Holistic filter pruning

In this section, a pruning loss is provided that can be used for common DNN training to prune filters and neurons by gradient descent. The pruning rates are freely adjustable and automatically distributed over the individual layers. The pruning itself is induced via the channel-wise scaling factors of the batch-normalization layers considering the respective layer sizes. The training objective combines the learning task $\mathcal{L}_{\text{learning}}$ and the pruning task $\mathcal{L}_{\text{pruning}}$ such that both are solved simultaneously during training:

$$\mathcal{L} = \mathcal{L}_{\text{learning}} + \lambda \, \mathcal{L}_{\text{pruning}} . \quad (4)$$

Here, $\lambda$ is the pruning parameter that scales the weighting between both tasks.

### 4.1. Indicator function

As demonstrated in section 3, the pruning of complete filters can be done via the channel-wise scaling factors of the batch-normalization layers. Thus, we first implement a magnitude based indicator function that determines whether the absolute value of $\gamma$ is smaller than the magnitude $t$:

$$\Phi(\gamma, t) = \begin{cases} 0 & \text{if } |\gamma| \leq t \\ 1 & \text{if } |\gamma| > t \end{cases} . \quad (5)$$

If the indicator functions outputs zero, the respective channel is considered inactive and would be deleted after training. As can be seen in Figure 2, $\Phi$ is a non-smooth quantization function whose gradient is zero almost everywhere. Therefore, we utilize the straight-through estimator (STE, [12]) which is widely used in network quantization to approximate the local gradient of step-functions during backpropagation. However, since $\Phi$ is symmetrical to the y-axis (in contrast, fixed-point quantization functions are usually symmetrical to the origin), we flip the estimator on the y-axis as well:

$$\frac{\partial \Phi(\gamma)}{\partial \gamma} = \begin{cases} -1 & \text{if } \gamma \leq 0 \\ 1 & \text{if } \gamma > 0 \end{cases} . \quad (6)$$

As shown in figure 2, this is the most suitable approach for approximating $\Phi$ with linear segments. As a result, the gradient estimator is easy to implement and non-zero for any input value.

Liu *et al.* found that scaling factors with absolute values below $10^{-4}$ can be set to zero without a noticeable drop in accuracy [26]. Therefore, we use $t = 10^{-4}$ for our experiments. According to equation 3, this results in the channel output being approximately equal to the batch-normalization bias $\beta_{l,c}$,

$$\hat{a}_{l,c} = \hat{w}_{l,c} * x_{l-1} + \hat{b}_{l,c} \overset{|\gamma_{l,c}| < 10^{-4}}{\approx} \beta_{l,c}, \quad (7)$$

which is independent from the channel input. The bias is propagated through the following convolution or fully-connected layer and shifts the resulting feature maps. However, this shift is corrected by the following batch-normalization layer by subtracting the mean over the respective mini-batch. After training, both the scaling factor and the bias of the batch-normalization layers are set to zero if the indicator function outputs zero.

### 4.2. Pruning loss

The complexity of a DNN can be specified on the one hand by the number of parameters $P$ and on the other hand by the number of floating-point multiplications $M$ that are needed to propagate one sample through the network. If $\widetilde{P}$ and $\widetilde{M}$ denote the number of parameters and multiplications of the pruned model and $P^*$ and $M^*$ specify the desired target values, the deviation between the pruned model and the target size can be described by the following loss function:

$$\mathcal{L}_{\text{pruning}} = \text{relu}\left(\frac{\widetilde{P} - P^*}{P}\right) + \text{relu}\left(\frac{\widetilde{M} - M^*}{M}\right) \quad (8)$$

with $P$ and $M$ being the number of parameters and multiplications of the original model. The terms within the rectifier functions denote the normalized differences between the current and the desired mode size with $1 - P^*/P$ being the desired pruning rate and $1 - \widetilde{P}/P$ being the current pruning rate. Thus, both summands vary between zero and the desired pruning rates. For example, if the goal is to prune 50% of the parameters and 40% of the required multiplications, the pruning loss takes values between 0 and 0.9.

Both the original and the desired network sizes are constant values: the former is fixed whereas the latter is specified by the user. Therefore, $\widetilde{P}$ and $\widetilde{M}$ remain the only variable quantities in equation 8. Utilizing the indicator function from equation 6, the number of parameters in a feed-forward neural network can be calculated as follows:

$$\widetilde{P} = \sum_{l=1}^{L-1} P_l \underbrace{\left(\frac{1}{C_{l-1} C_l} \sum_{c=1}^{C_{l-1}} \Phi(\gamma_{l-1,c}) \sum_{c=1}^{C_l} \Phi(\gamma_{l,c})\right)}_{\text{pruning rate of intermediate layers}}$$

$$+ P_L \underbrace{\left(\frac{1}{C_{L-1}} \sum_{c=1}^{C_{L-1}} \Phi(\gamma_{L-1,c})\right)}_{\text{pruning rate of the last layer}} \quad (9)$$

Here, $l$ denotes the layer index, $L$ the number of layers, $C_l$ the number of channels in layer $l$, and $P_l$ the original number of weights in layer $l$. The terms within the brackets correspond to the pruning rates of the respective layer and depend on the balance between active and inactive channels. The pruning ratios are scaled with the respective channel sizes and added together over the number of layers. The same calculation can be done analogously for the number of pruned multiplications:

$$\widetilde{M} = \sum_{l=1}^{L-1} M_l \underbrace{\left(\frac{1}{C_{l-1} C_l} \sum_{c=1}^{C_{l-1}} \Phi(\gamma_{l-1,c}) \sum_{c=1}^{C_l} \Phi(\gamma_{l,c})\right)}_{\text{pruning ratio in intermediate layer } l}$$

$$+ M_L \underbrace{\frac{1}{C_{L-1}} \sum_{c=1}^{C_{L-1}} \Phi(\gamma_{L-1,c})}_{\text{pruning ratio in last layer } L} . \quad (10)$$

Hence, during each forward pass the pruning loss calculates the deviation between the current and the desired model size in terms of the number of parameters and multiplications. The gradients can be backpropagated by utilizing the gradient estimator of the indicator function.

### 4.3. Shortcut connections

State-of-the-art DNN architectures such as ResNet [7], DenseNet [14], or MobileNet use shortcut connections between layers which add the output feature maps of the layers. This makes filter pruning more complicated since shortcut connections can reactive already pruned channels. Several solutions have been proposed for this problem: In [22, 28] layers with shortcut connections were not pruned to avoid the problem of reactivated channels. However, skipping the layers with shortcut connections redudces the feasible pruning ratio. Furthermore, in [26, 11] feature maps were sampled in front of each residual block to reduce their dimension. Yet, sampling layers bring additional computation cost. The authors of [34] proposed a group pruning method in which layers connected by a shortcut connection share the same pruning patterns.

In our case, the application of shortcut connections is not a problem as long as the counting functions from equation 9 and 10 are implemented correctly. Consequently, when calculating the layer-wise pruning rates, it must be taken into account whether a shortcut connection is added and if so, whether the inactive channels match on both sides. This can be done by using a mask that consists of the element-wise sums of the absolute values of the batch-normalization scaling factors. The mask is then processed by the indicator function $\Phi$ to calculate the pruning rates.

### 4.4. Implementation details

Algorithm 4.4 shows how a DNN can be pruned using *Holistic Filter Pruning* (HFP).

**Sparsity learning**: After each forward-pass, the pruning loss is calculated according to equation 8 and added to the learning loss. Subsequently, the parameters are updated using SGD optimization with a nesterov momentum of $0.9$. We train until the number of given epochs is reached.

**Regularization parameter**: In equation 8, the pruning parameter $\lambda$ regularizes the weighting between the learning task on the one hand and the pruning loss on the other. Hence, $\lambda$ should be chosen such that both losses are in the same order of magnitude. Therefore, we define $\lambda$ such that $\lambda \mathcal{L}_{\text{pruning}}$ is equal to the expectation value of the learning loss over the training set. E.g., if the average cross-entropy loss for an untrained model is $7.25$ on ImageNet, and the desired pruning rate is $0.5$ for both parameters and multiplications, $\lambda$ is equal to $7.25$. Furthermore, since we use pre-trained models, we recommend heating up the pruning parameter from one over the training epochs.

**Fine-tuning**: After training, channels whose scaling factors are set to zero by the indicator function are completely deleted from the network architecture. Subsequently, the remaining architecture is retrained for three epochs to update the batch-statistics of the batch-normalization layers.

---

**Algorithm 1** The procedure to prune a DNN with *Holistic Filter Pruning*. The steps that have to be implemented are in line 7 and 12.

1: **Input**: Pre-trained model $O$, Training Data $(X, Y)$, Learning task $\mathcal{L}_{\text{learning}}$, Target size $\{P^*, M^*\}$.
2: model $\leftarrow O$
3: **for** e in epochs **do**
4:     **for** (data, target) in $(X, Y)$ **do**
5:         out  = model(data)
6:         loss  = $\mathcal{L}_{\text{learning}}$(out, target)
7:         loss += $\lambda \mathcal{L}_{\text{pruning}}$(model, $P^*, M^*$)    Sparsity learning
8:         loss.backward( )
9:         SGD.step(model)
10:     **end for**
11: **end for**
12: model $\leftarrow$ Prune(model)
13: model $\leftarrow$ Retrain(model)    ▷ Fine-tune for 3 epochs
14: **return** model

---

## 5. Experiments

In this section, we evaluate *Holistic Filter Pruning* (HFP) on common benchmark data-sets including CIFAR-10 and ImageNet. First we compare with state-of-the-art filter pruning methods before giving insights into the training procedure of HFP. The baselines of experiments on CIFAR-10 are calculated by training for 150 epochs using SGD optimization with the nesterov momentum set to 0.9 and a batch-size of 64. The learning rate is reduced linearly during the training from $10^{-2}$ to $10^{-4}$. For ImageNet, the baselines are taken from the torchvision model zoo[1].

### 5.1. VGG-8 and ResNet-56 on CIFAR-10

CIFAR-10 is an image classification task with 10 different classes [17]. The data consists of $32 \times 32$ color images and is divided into 50,000 training and 10,000 test samples. We preprocess the images as recommended in [14] and use a batch-size of 64. Furthermore, we train for 150 epochs and linearly decrease the learning rate from 0.02 to $10^{-4}$.

Table 1 shows the pruning results of VGG-8. We specify to prune the number of parameters by 90% and the number of multiplications by 80%. Thus, we achieve comparable pruning rates to *HRank* but outperform the accuracy significantly by 3%. In comparison to Zhao *et al.* and *SSS*,

---

[1]https://pytorch.org/docs/stable/torchvision/models.html

### VGG-8 on CIFAR-10

Table 1. Top-1 accuracy and percentage reduction in the number of multiplications and parameters.

| Method | Flops % ↓ | Params % ↓ | Top-1 % |
|---|---|---|---|
| Baseline | - | - | 94.89 |
| SSS [16] | 41.6 | 73.8 | 93.02 |
| Zhao *et al*. [36] | 39.1 | 73.3 | 93.18 |
| GAL-0.1 [25] | 45.2 | 82.2 | 90.73 |
| HRank [24] | 65.3 | 82.1 | 92.34 |
| HRank [24] | 76.5 | 92.0 | 91.23 |
| **HFP** | 82.0 | 90.0 | **94.21** |

### ResNet-56 on CIFAR-10

Table 2. Top-1 accuracy and percentage reduction in the number of multiplications and parameters. Results marked with '-' are not reported by the authors.

| Method | Flops % ↓ | Params % ↓ | Top-1 % |
|---|---|---|---|
| Baseline | - | - | 93.30 |
| NISP [35] | 35.50 | 42.40 | 93.01 |
| DCP [38] | 47.10 | 70.30 | 93.79 |
| CP [11] | 50.00 | - | 91.80 |
| FPGM [10] | 52.60 | - | 93.26 |
| GBN-40 [34] | 60.10 | 53.50 | 93.41 |
| GBN-60 [34] | 70.30 | 66.70 | 93.07 |
| HRank [24] | 50.00 | 42.40 | 93.17 |
| HRank [24] | 74.10 | 68.10 | 90.72 |
| **HFP** | 56.00 | 50.00 | **93.30** |
| **HFP** | 76.09 | 71.58 | **92.31** |

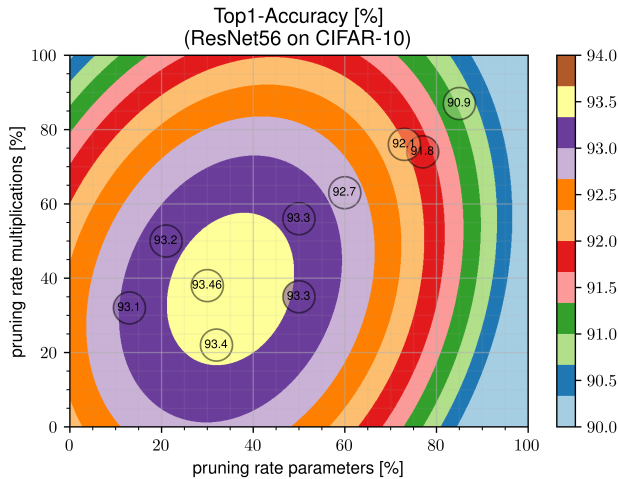

Figure 3. Top-1 accuracies of ResNet-56 on CIFAR-10 with different pruning rates. The performance values are illustrated by colored level curves created by fitting a second-order polynomial.

### ResNet-50 on ImageNet

Table 3. Labels have the same meaning as in Table 2.

| Method | Flops % ↓ | Params % ↓ | Top-1 % |
|---|---|---|---|
| Baseline | - | - | 76.15 |
| *NIPS 2018, NIPS 2019, CVPR 2019* | | | |
| DCP [38] | 55.76 | 51.45 | 74.95 |
| FPGM [10] | 53.50 | - | 74.83 |
| GBN-60 [34] | 40.54 | 31.83 | 76.19 |
| GBN-50 [34] | 55.06 | 53.40 | 75.18 |
| *CVPR 2020* | | | |
| Hinge [23] | 53.45 | - | 74.70 |
| He *et al*. [8] | 60.80 | - | 74.56 |
| DMCP [4] | 73.17 | - | 74.40 |
| HRank [24] | 62.10 | - | 71.98 |
| HRank [24] | 76.04 | - | 69.10 |
| **HFP** | 60.25 | 41.01 | **76.08** |
| **HFP** | 69.80 | 53.83 | **75.36** |
| **HFP** | 73.45 | 59.20 | **74.81** |
| **HFP** | 78.24 | 68.48 | **74.14** |

we achieve higher pruning rates, while simultaneously increasing the accuracy by more than 1%. Compared to the baseline accuracy, we are able to reduce the number of parameters by 90% with an accuracy drop of 0.6%.

Table 2 shows the pruning results on the ResNet-56 architecture. We use two different settings with target reductions of 50% and 70%, respectively. Thus, we are able to prune both the parameters and the multiplications by at least 50% with no loss of accuracy. In comparison to *HRank*, we achieve higher pruning rates with a slightly improved Top-1 accuracy. To the best of our knowledge, we are the first to reduce the number of multiplications by more than 75% while at the same time reducing accuracy by less than 1.5%. *GBN* achieves a slightly higher Top-1 accuracy for comparable pruning rates. Additionally, figure 3 illustrates the level curves of various experiments with different pruning rates on ResNet-56. One can observe that pruning the parameters has a greater impact on the performance than pruning the multiplications.

### 5.2. ResNet-50 and ResNet-18 on ImageNet

ImageNet is an image classification task which provides 1000 different class labels. We use the data from 2012 (ILSVRC12 [18]) which consists of 1,281,167 training and 50,000 test samples. We preprocess the data by subtracting the mean and dividing by the standard-deviation over the training set. For data-augmentation we apply random horizontal flips and crop the images to $224 \times 224$. We train for 100 epochs with a batch-size of 256 and linearly decrease

### ResNet-18 on ImageNet

Table 4. Top-1 accuracy and percentage reduction in the number of multiplications and parameters. Results marked with '-' are not reported by the authors.

| Method | Flops % ↓ | Params % ↓ | Top-1 % |
|---|---|---|---|
| Baseline | - | - | 69.75 |
| SFP [9] | 41.80 | - | 67.10 |
| FPGM [10] | 41.80 | - | 68.41 |
| **HFP** | 36.30 | 22.07 | **69.15** |
| **HFP** | 45.00 | 37.27 | **68.53** |

### ResNet-50 on ImageNet

Table 5. Top-1 accuracy and percentage reduction in the number of multiplications and parameters for different values of $\lambda$.

| $\lambda$ | Flops % ↓ | Params % ↓ | Top-1 % |
|---|---|---|---|
| 1. | 48 | 36 | **76.41** |
| 7.25 | 62 | 42 | **75.73** |
| $1 \rightarrow 7.25$ | 60 | 41 | **76.08** |

the learning rate from $10^{-1}$ to $10^{-4}$.

Table 3 shows the pruning results of ResNet-50. To enable accurate comparisons, we evaluate four configurations with various pruning rates and compare with the latest pruning results from CVPR2020. The first configuration reduced the number of multiplications by 60% with no significant loss in the accuracy. The second configuration achieves both higher pruning rates and higher accuracy than in [34, 23, 8]. The third configuration yields a reduced number of multiplications and slightly improved accuracy in comparison with [4]. Furthermore, HFP is able to reduce the number of multiplications by 78% with only 2% loss in the accuracy.

Table 4 shows the pruning results of ResNet-18. ResNet-18 is much smaller than ResNet-50, less over-parameterized and consequently more difficult to prune. HFP provides new state-of-the-art performance with 36% reduced multiplications and only 0.6% accuracy decrease. The second configuration reduces the number of multiplications by 45% with only 1.2% loss in the accuracy

### 5.3. Ablation study on the pruning parameter

Table 5 shows the pruning results of ResNet-50 with the aim of pruning 60% of the multiplications and 40% of the parameters by using different values of the pruning parameter $\lambda$. The first experiment uses the constant value $\lambda = 1$. As noticeable, the desired pruning rates are not reached since the weighting of the pruning loss is to low. The second experiment uses $\lambda = 7.25$ which results from the consideration in section 4.4. Indeed, the desired pruning rates are fulfilled. However, the accuracy drops below 76% since the imbalance is high at the beginning of the training. The third experiment utilizes the proposed strategy of heating up $\lambda$ from 1 to 7.25: The pruning rates are still fulfilled and the accuracy increases in comparison to the second experiment.

### 5.4. Pruning rate allocation of VGG-8

The distribution of the overall pruning budget to the individual layers is a well-known problem in filter pruning. HFP automatically distributes the pruning rates among the individual layers such that the pruning loss is minimized.

VGG-8 consists of six convolution layers and two fully-connected layers. The convolution layers are especially expensive regarding the number of multiplication whereas the first fully-connected layer owns most of the parameters. Thus, we analyze two experiments: a) with the aim of pruning 90% of the parameters and b) with the aim of pruning 90% of the multiplications. Figure 4 shows the layer-wise pruning rates for both experiments as well as the proportional layer sizes regarding the number of parameters and multiplications. In the first experiment, HFP primarily reduces the layers which contribute most to the number of parameters (conv6 and fc7). Especially fc7 has a large number of parameters and is therefore pruned by approximately 97%. In contrast, the second experiment mainly leads to a reduction of the convolution layers as they offer more potential for saving multiplications. Consequently, we can observe that HFP distributes flexible pruning rates over the individual layers. Furthermore, the distribution of the pruning budget varies depending on the target reduction. Comparisons with the layer sizes regarding the number of multiplications and parameters result in a meaningful distribution.

### 5.5. Visualization of ResNet-56

This section analyzes how the overall reduction of parameters and multiplications is proportionally distributed among the individual layers. For example, if 1000 parameters are pruned and the first layer is reduced by 150 parameters, then the proportional contribution of the first layer to the parameter pruning is 15%. Figure 5 shows the proportional pruning rates of ResNet-56 with 56% pruned multiplications and 50% pruned parameters (section 5.1, table 1). The pruning rates are shown for different training epochs and refer to the pruning result at that time step (e.g., after 10 epochs 47% of the multiplications were reduced). The first diagram shows the proportional pruning rates for the multiplications while the second diagram shows the proportional pruning rates for the parameters. Additionally, the diagrams show the total number of multiplications and parameters of the unpruned layers (dotted lines). In both figures, the three basic blocks of the ResNet architecture are visible and marked with A, B, and C. In case of the pruned multiplications, the proportional pruning rates of the individual layers change over the epochs. While in block A the proportional pruning rates decrease as training progresses, the rates in block C increase: the allocation of the pruning budget changes continuously during training. In the end, the second and third block achieve slightly higher pruning rates compared to the first block. However, the differences are comparably small as all intermediate layers share the same number of multiplications. In contrast, the proportional pruning rates of the parameters change significantly depending on the layer index. With an increasing layer size, the pruning rates also increase. Since block C has the highest contribution to the total number of parameters, it also shows the highest contribution to the pruning rate.

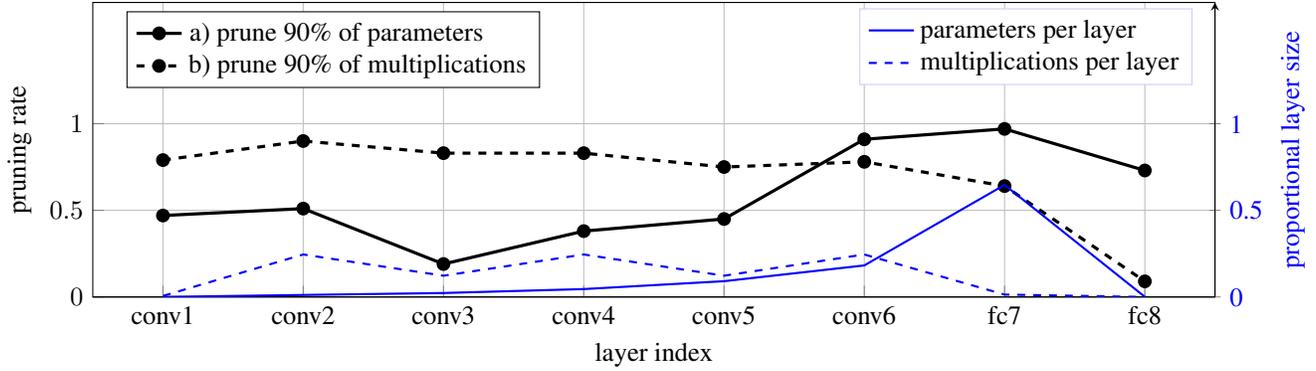

Figure 4. Pruning rates of all layers in VGG-8 for two different experiments: a) with the aim of pruning 90% of the parameters and b) with the aim of pruning 90% of the multiplications. Depending on the target reduction, the pruning budget is distributed differently over the individual layers: a) reduces layers with many parameters while b) especially prunes the convolution layers with many multiplication.

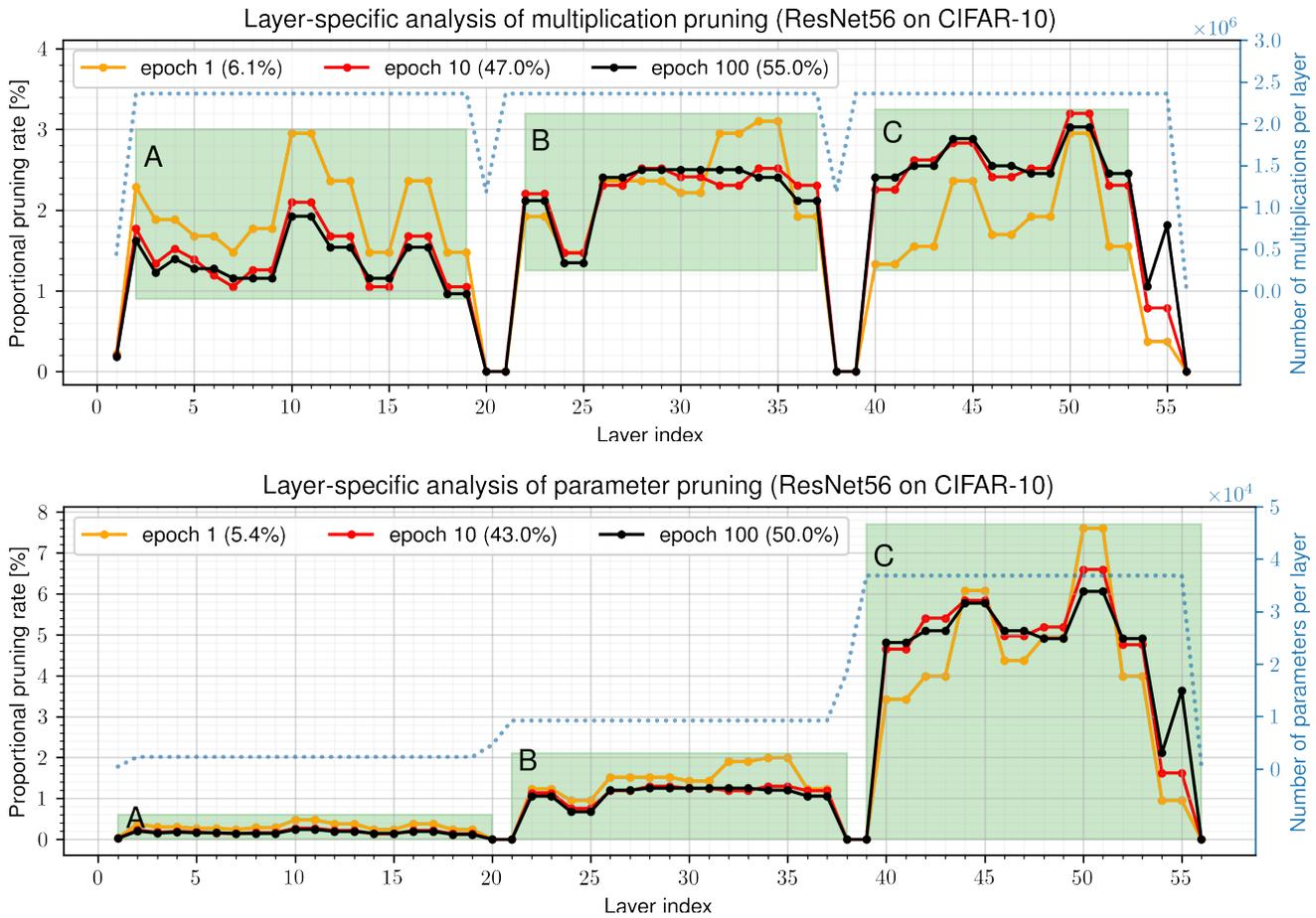

Figure 5. The upper plot shows the proportional pruning rates of the individual layers of ResNet-56 (with 56% reduced multiplications and 50% reduced parameters, table 1). Proportional pruning rates indicate the contribution of single layers to the overall pruning rate. E.g., if 1000 multiplications are pruned from the model and the first layer is reduced by 150 multiplications, the proportional pruning rate of the first layer is 15%. The lower plot indicates the proportional pruning rates for the number of parameters.

## 6. Conclusion

We propose *Holistic Filter Pruning* (HFP), a simple and powerful filter pruning method to reduce the complexity of trained DNNs. HFP uses a pruning loss that takes accurate pruning rates for the number of both parameters and multiplications into account. After each forward pass, the deviation between the current model size and the target size is calculated. By gradient descent, the pruning rates are distributed over the individual layers such that the target size is fulfilled. The loss function fits seamlessly into the training of DNNs and uses the channel-wise scaling factors of the batch-normalization layers to calculate the model size. Thus, no additional variables need to be defined and the implementation effort is low. Especially for large pruning rates ($> 70\%$), HFP yields excellent performance and outperforms recent approaches by up to 5%.